\documentclass[runningheads]{llncs}

\usepackage[T1]{fontenc}
\usepackage{amsmath}
\usepackage{amssymb}
\usepackage{xcolor}
\usepackage{booktabs}
\usepackage{hyperref}

\usepackage{graphicx}
\usepackage{graphbox}
\usepackage{subcaption} 
\usepackage{float}
\floatstyle{plain}  % 基础样式（可改为"boxed"等）
\restylefloat{figure}  % 应用于图片
\restylefloat{table}   % 应用于表格
\begin{document}
\title{FlexMUSE: Multimodal Unification and Semantics Enhancement Framework with Flexible interaction for Creative Writing}

\titlerunning{FlexMUSE: Multimodal Unification and Semantics Enhancement Framework with Flexible interaction for Creative Writing}

\author{Jiahao Chen *\inst{1} \and
Zhiyong Ma *\inst{1,2} \and Wenbiao Du \inst{1} \and Qingyuan Chuai \dag\inst{1}}
% \cortext[mycorrespondingauthor]{Corresponding author}
\institute{
Cao Tu Li(Guangzhou) Technology Co., Ltd, China
\and
South China University of Technology, China
\\
\email{\dag chuaiqingyuan@aiiiin.com}}
\maketitle              % typeset the header of the contribution
\begin{abstract}
%研究背景 2
Multi-modal creative writing (MMCW) aims to produce illustrated articles.
% 说明这个任务的独特性 
Unlike common multi-modal generative (MMG) tasks such as storytelling or caption generation, MMCW is an entirely new and more abstract challenge where textual and visual contexts are not strictly related to each other.
%现有挑战 2 （四大问题：不灵活、难训练、模态语义不一致）
Existing methods for related tasks can be forcibly migrated to this track, but they require specific modality inputs or costly training, and often suffer from semantic inconsistencies between modalities.
Therefore, the main challenge lies in economically performing MMCW with flexible interactive patterns, where the semantics between the modalities of the output are more aligned.
%现有解决思路 2
% To tackle this challenge, previous work manually combine the responses of online language models (LMs) or straightforwardly merge modalities' information into LMs through end-to-end training.
% The former heavily relies on rigid interactions and online service, while the latter is data-driven but sensitive to the quality and quantity of data and requires deeper mining for semantic knowledge.
% 现有挑战1：缺数据集
% In addition, calibrated corpuses for MMCW with text-image pairs are rare in the Chinese and even in the English community. 
% Moreover, calibrated corpuses for MMCW are rare in Chinese community. 
%研究内容概述 4
% In this work, we propose FlexMUSE with a T2I module to achieve optional visual input, emphasizing the unification between modalities by the modality semantic alignment gating (msaGate) and the modality consistency Direct Preference Optimization (mscDPO), while aiming to enhance the semantic relevance by cross-modality fusion.
% The msaGate restricts the general semantic clues via probabilistically masking the textual input.
% The mscDPO extends the rejected samples for more diverse and finer distinctions through supervision.
In this work, we propose FlexMUSE with a T2I module to enable optional visual input.
FlexMUSE promotes creativity and emphasizes the unification between modalities by proposing the modality semantic alignment gating (msaGate) to restrict the textual input.
Besides, an attention-based cross-modality fusion is proposed to augment the input features for semantic enhancement.
The modality semantic creative direct preference optimization (mscDPO) within FlexMUSE is designed by extending the rejected samples to facilitate the writing creativity.
%所得结论 2
% To advance the field of MMCW, we expose a Chinese dataset called ArtMUSE, which contains calibrated text-image pairs.
Moreover, to advance the MMCW, we expose a dataset called ArtMUSE which contains with around 3k calibrated text-image pairs.
FlexMUSE achieves promising results, demonstrating its consistency, creativity and coherence.

\end{abstract}
\section{Introduction}
\begin{figure}[htbp]
    \centering
    \begin{subfigure}[b]{0.55\textwidth}
        \centering
        \includegraphics[width=1\linewidth]{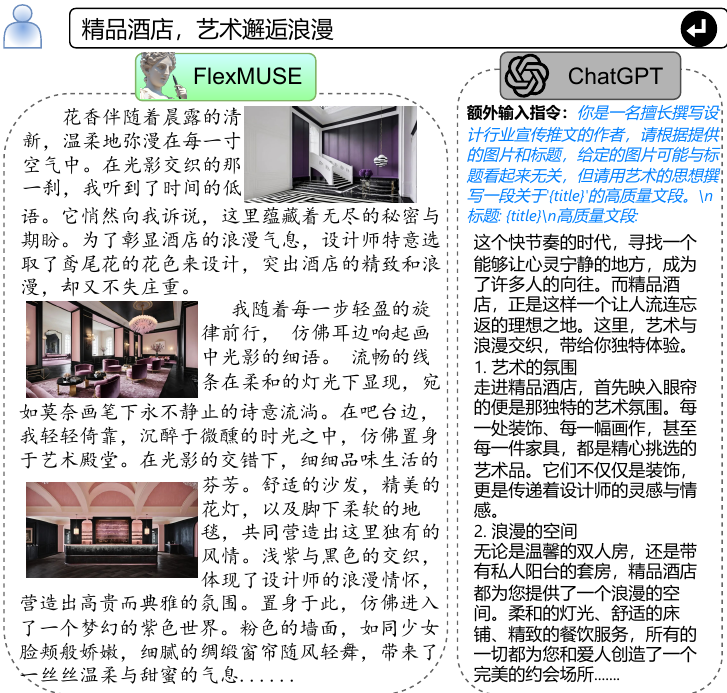}
        \caption{
        % The case comparison between two methods given same Chinese topic (Boutique hotels, where art meets romance).
        % For fairness, we provide well-designed instructions only for online LLMs like ChatGPT (in blue).
        % Intuitively, FlexMUSE achieves more promising semantic consistency and artistic creativity, where the first paragraph echoes the shadows and hotel rooms within the figure, while the second paragraph describes visual elements such as couches and carpets with vivid writing style.
        }
        \label{fig:case_study}
    \end{subfigure}
    \hspace{0.01\textwidth}  % 减小子图间距
    \begin{subfigure}[b]{0.41\textwidth}
        \centering
        \includegraphics[width=\linewidth]{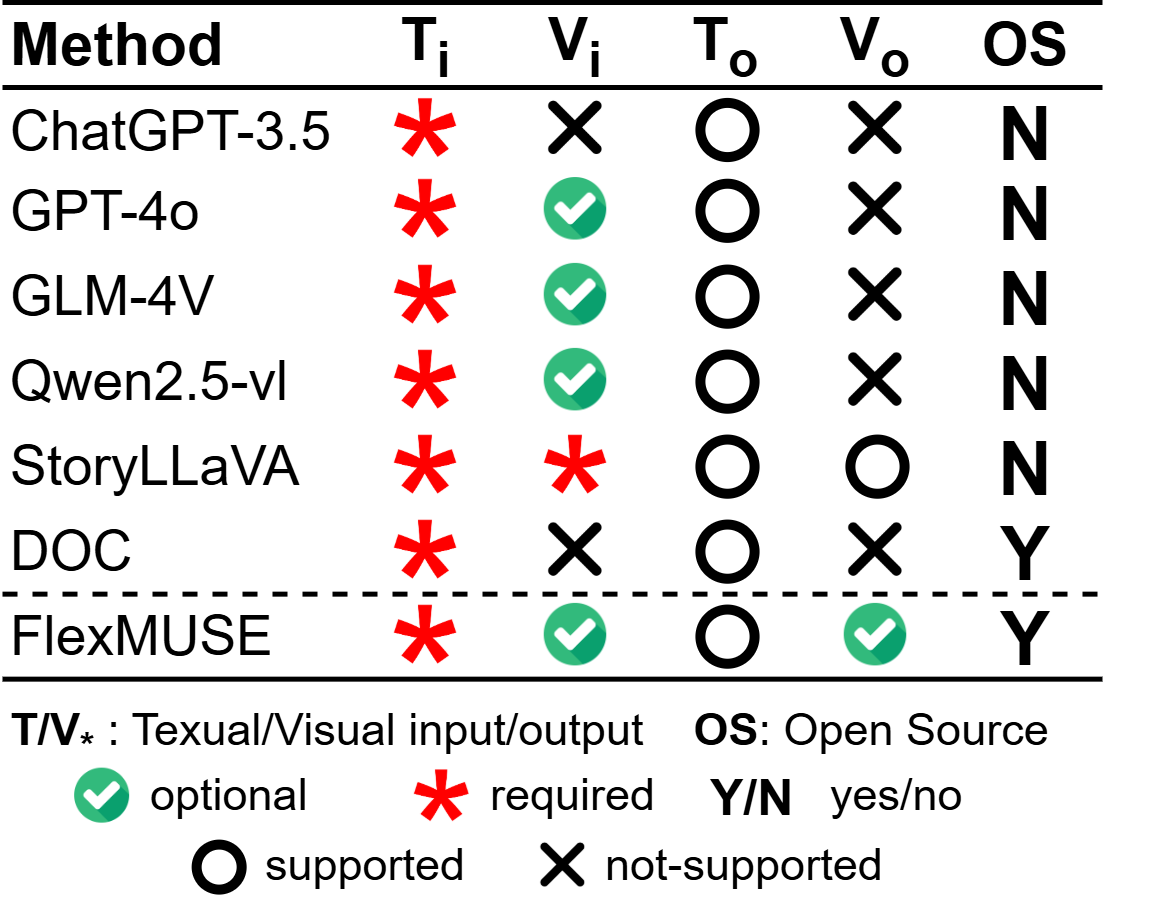}
        \vspace{0.8cm}
        \caption{}
        \label{fig:interaction}
    \end{subfigure}
    
    % \caption{
    %     The sub-figure (\ref{fig:case_study}) shown the comparison between two methods given same Chinese topic (Boutique hotels, where art meets romance).
    %     For fairness, we provide well-designed instructions only for online LLMs like ChatGPT (in blue).
    %     Intuitively, FlexMUSE achieves more promising semantic consistency and artistic creativity, where the first paragraph echoes the shadows and hotel rooms within the figure, while the second paragraph describes visual elements such as couches and carpets with vivid writing style.
    %     The sub-figure (\ref{fig:interaction}) show the comparison about interaction and open source (OS) status. 
    %     T/V represent the textual/visual pattern, and subscripts i/o represent the input/output, respectively.
    %     }

        \caption{
        (a) comparison between two methods given same Chinese topic (Boutique hotels, where art meets romance).
        For fairness, we provide well-designed instructions only for online LLMs like ChatGPT (in blue).
        Intuitively, FlexMUSE achieves more promising semantic consistency and artistic creativity, where the first paragraph echoes the shadows and hotel rooms within the figure, while the second paragraph describes visual elements such as couches and carpets with vivid writing style.
        (b) comparison in interaction and open source (OS) status. 
        % T/V represent the textual/visual pattern, and subscripts i/o represent the input/output, respectively.
        }
    \label{fig:combined}
\end{figure}

% MMCW背景，有什么意义
% 图文并茂的推文和帖子是公认的数字艺术，并十分适合展现作者高雅且抽象的设计，尤其在设计和广告领域。
The digital illustrated articles (e.g. Tweets or posts) are recognized as modern MUSE (art) and well suited to exhibit the elegant and abstract idea of the author, especially in the field of architecture, design, and advertising \cite{Bhattacharyya_Palmer_Heckman_2024}.
% MMCW是做什么的
Multi-modal creative writing (MMCW) is supposed to generate these articles by given pieces of information like topic or images or both.
% 说明MMCW属于一个新的task
Unlike common multi-modal generation (MMG) \cite{Yang_Klein_Peng_Tian_2023,Zang_Tang_Zhang_Zhao_Lv_Pei_Liang_2024}, MMCW belongs to a new track which is more abstract that the textual and visual context are not strictly related to each other.
The automation of MMCW brings
a challenge at the intersection of AGI 
\cite{bai2025qwen25vltechnicalreport,Wu_Tang_Wang_Zeng_Li_Tong_2024} and cognitive science, offering
transformative potential for both augmenting human creativity and
multi-modal generation \cite{Wang_Yasunaga_Ren_Wada_Leskovec_2023}, as shown in Fig. \ref{fig:case_study}. 
% \begin{figure}[htbp]
%     \centering
%     % \includegraphics[width=1\linewidth]{img/case_study.pdf}
%     \includegraphics[width=0.55\linewidth]{img/case_study.pdf}
%     \caption{
%     The case comparison between two methods given same Chinese topic (Boutique hotels, where art meets romance).
%     For fairness, we provide well-designed instructions only for online LLMs like ChatGPT (in blue).
%     Intuitively, FlexMUSE achieves more promising semantic consistency and artistic creativity, where the first paragraph echoes the shadows and hotel rooms within the figure, while the second paragraph describes visual elements such as couches and carpets with vivid writing style.
%     }
%     \label{fig:case_study}
% \end{figure}

% 困难1和2--不灵活、难训练
Existing methods for MMG \cite{Yang_Ge_Li_Chen_Ge_Shan_Chen_2024} can forcibly transfer to this track through manually combining the responses of online language models (LMs), or merge modalities' information via end-to-end training LMs, even large language models (LLMs) and multi-modal large language models (MLLMs) \cite{Yang_Xiao_Huang_Zhong_2025}.
% Some existing literature \cite{Yang_Ge_Li_Chen_Ge_Shan_Chen_2024} suppose to enter specific modalities to online language models (LMs) and manual combine the responses.
% While some methods train LMs even large language models (LLMs) or multi-modal large language models (MLLMs) via end-to-end manner \cite{Yang_Xiao_Huang_Zhong_2025}. 
% 解释困难1和2所带来的缺陷
% 前者受困于僵硬的交互和在线服务的知识，这与MMCW节省人力资源和保护创意的初衷背道而驰。
% \begin{figure}[htbp]
%     \centering
%     % \includegraphics[width=1\linewidth]{img/interaction.png}
%     \includegraphics[width=0.55\linewidth]{img/interaction.png}
    
%     \caption{
%     The comparison about interaction and open source (OS) status. 
%     T/V represent the textual/visual pattern, and subscripts i/o represent the input/output, respectively. 
%     }
%     \label{fig:interaction}
% \end{figure}
The former not only trapped in the rigid interaction but also relies on online services, 
as shown in Fig. \ref{fig:interaction}, 
which is opposite to the original intention of automated MMCW that saves human resources and protects ideas.
% 通过端对端的模式训练一个任务相关的语言模型来融合不同模态的信息是一种直观的手段，但存在前提条件，如需准备高性能计算硬件、充分且优质的数据、开源且易训练的基座，引申出困难3--模态语义不一致和困难4--缺数据
For the latter, training task-specific model to merge modalities' information is intuitive \cite{Mo_Liu_2024}, but comes with prerequisites such as the high-performance-computing hardwares, the well-prepared corpus, and the open-source backbones, which might be costly and easily suffer from semantic inconsistencies between modalities.
Moreover, semantic alignment between modalities, creativity, and coherence about generation are quite significant for MUSE.
%引出核心观点
%总结上述的所提的不足：对于MMCW，我们渴望寻求一个易于训练的模型并得到语义对齐更高，更具创意，更连贯的输出结果
% Thus, it is valuable to design a method with flexible interaction which can produce articles with better semantic alignment, creativity and coherent.
Then a natural question is: {\it How to efficiently construct this well-behaved method with flexible interaction for MMCW?}

% 简单铺垫框架
% 为保障文本生成能力，MMCW的框架通常需要使用LMs以接收用户的中心主旨并生成文本。
In general, methods tailored for MMG should leverage pre-trained models to extract and understand the input text \cite{GLM_2024,Raffel_Shazeer_Roberts_Lee_Narang_Matena_Zhou_Li_Liu_2023}.
% 由于图文并茂且语义相关内容的生成内容显然是更受人喜欢的，因此MMCW需要引入图像到输入和输出工作流程中。
% Meanwhile, since vivid and semantically relevant articles are more popular, it is reasonable for MMCW method to receive or generate images acting as the visual input anchors of system and output illustration.
Meanwhile, these methods should also receive or generate images acting as visual input anchors of system for generating illustrations.
% 为保障生成内容的语义关联性，视觉和文本的信息需要融合并输入到LMs中指导生成过程。
To ensure the semantic relevance of the visual and textual result, the information of two modalities are needed to fuse and inject into downstream modules \cite{Xue_Qian_Fang_Xu_2022,Wang_Zhao_Chen_Chen_Zheng_Shen_2024}.
% 然而，从实践角度来看，生成的文本通常会出现图文语义不一致的情况
However, the generated articles easily get trapped by semantic inconsistency between modalities and creativity deficit in MMCW practices.
% 另外，若LMs过分关注某些特定信息或缺乏多样化的监督信号，生成的内容将陷入创造性缺失的危险。
To tackle the aforementioned challenges, we propose FlexMUSE with several modules for chasing different properties.

% 介绍方法的模块
% 在用户输入标题或主题的前提下，通过为Writer引入T2I可以使框架的交互变得灵活
For flexible interactions (optional visual inputs) and vivid result, FlexMUSE equipped with diffusion model \cite{Rombach_Blattmann_Lorenz_Esser_Ommer_2022} as the text-to-image (T2I) modules, transferring the textual semantic to relevant but diversity figures.
% Each figure acts as visual anchor and is used as an illustration in the generated output.
% 从信息论的角度分析，为文本生成模型引入视觉信息后，系统的熵值将变大，导致模型的不稳定性变强，从而引发不同模态之间的语义不一致现象，使图文语义不相关。
% Inspired by the literature based on information theory \cite{Hu_Guo_Yu_Liu_2010,yadkori2024believebelievellm,zhang2023calibratingmultimodallearning}, when additional information is introduced, the entropy of the system might increase, leading to more significant instability. 
% We take this instability as a double-edged sword to trade-off between semantic consistency and creativity.

% 考虑到潜在的，由输入文本和图像所引起的语义冲突风险和过分关注文本线索导致结果多样性缺失问题，我们提出了一个简单但有效的设计，名为msaGate。
To alleviate the semantic conflicts between modalities, we propose a simple but effective gate design named modalities semantic alignment gate (msaGate).
% 解释设计的思路
% msagGate基于模态之间的语义相似性概率地掩码文本输入。
It masks textual inputs based on probability and the semantic similarity between modalities, filtering out redundant information while reducing the risk of increasing the information entropy \cite{Hu_Guo_Yu_Liu_2010,yadkori2024believebelievellm}.
% This seems counterintuitive but reasonable, since the visual meaning is close to but not exactly the same as the semantic of textual input, either the generated or input image.
% 我们利用这些细粒度的差异来保障生成的创造性
% Since the meanings of images are close but not identical to the semantics of textual cue, we try to leverage these differences to lead to more creative generation.
% 补充关于融合与增强的动机
% 不同输入模态的语义必然存在近似和差异
% 求同存异地利用这些知识将有助于提高生存内容的质量
% Since there must be similarities and differences in the semantics of heterogeneous input modalities, we claim that seeking common ground while preserving the diversity of input knowledge is beneficial for the approach to construct MUSE.
Besides, it is intuitively reasonable that seeking common while preserving diversity of input knowledge is beneficial for mitigating creativity deficit in MMCW.
% 为了增强LMs输入的语义信息，FlexMUSE聚焦于模态间的语义交互并进行深入挖掘，提出了一张基于注意力机制的夸模态融合模组
Based on this view, we propose a cross-modality fusion module which attempts to capture the correlation between vision and text by attention mechanism, then leverage this correlation to conduct semantic augmentation in latent space. 

On the other hand, unlike other MMG tasks, MMCW need to consider the topic unification over paragraphs for semantic consistency, and retain discrepancy in content for creativity, which is an abstractive object and difficult to achieve by directly training LMs with SFT manner in practices.
% 为了减缓监督训练带来的语义不一致退化，我们引入并改造了DPO，提出了mscDPO。
On this purpose, we simulate human thoughts and draw on the human preference-aligned reinforcement learning (RLHF) \cite{Rafailov_Sharma_Mitchell_Ermon_Manning_Finn_2024,openai2024gpt4ocard}, proposing the modality semantic creative direct preference optimization, named mscDPO, by extending the sampling method for rejected samples from conventional DPO.
    
% 为了推动领域，我们公开数据集
% 引1篇这个多模态生成领域的文章，再引一篇综述
To advance the MMG \cite{Shahriar_2022}, we release a manually calibrated dataset on topics related to architectural art namely ArtMUSE, collated from Chinese social media, with approximately 3k text-image pairs.

Experimental results show the superiority of FlexMUSE within several aspects, demonstrating that it outperforms state-of-the-art methods across both automated metrics and LLM-based evaluations, with at least 9.091\% improvement in BertScore, 12.27\% in creativity, and  2.61\%  in coherence.
Ablation studies further validate the effectiveness of the purposed components. 
Moreover, FlexMUSE shows friendly computation cost with low VRAM usage.
Additional tests on hyperparameter robustness further indicate that its performance remains stable despite variations in hyperparameter values within a certain range.

% 实验情况可以直接copy到后面
% 引两篇用了这几个指标的文章
% Finally, we evaluate FlexMUSE on ArtMUSE, showing promising performance in CIDEr, MENTOR
% BLEU, and ROUGE scores \cite{Chenjiahao, Chenjiahao}.
% Besides, we verify the effectiveness of FlexMUSE in mitigating semantic inconsistencies and it exhibit decent in consistency, creativity, and coherence.
% 此处添加实验情况，在什么数据集上跑出什么效果，指标上比别人牛逼在哪

% 总结贡献点
Our contributions are concluded as follows:
\begin{enumerate}
    % 框架
    \item We propose FlexMUSE, a flexible and efficient MMG framework equipped with the msaGate, the cross-modality fusion module, and the mscDPO. 

    % 提出数据集
    \item We release ArtMUSE, a manually calibrated dataset with approximately 3k text-image pairs from Chinese social media for MMCW.
    
    % 实验结果不俗和并具有可拓展性
    \item Experimental results of FlexMUSE show promising performance in various criteria, verify its effectiveness in mitigating semantic inconsistencies, and validate its decent quality in creativity, and coherence.

\end{enumerate}

\section{Background}
% \subsection{Multi-modal Generation}
% -- 介绍多模态生成是干嘛的，有什么应用，有什么任务
% Multi-modal generation tasks integrate heterogeneous data modalities (e.g., text, image, and audio) to produce cross-modal outputs \cite{Yu_Chung_Yun_Kim_Kim_2021}. 
MMG aims to integrate heterogeneous data modalities to produce outputs \cite{Wehrmann_Kolling_Barros_2020}. 
It has gained attention in the community arising from mimicking human cognition, and has established various applications such as human-computer interaction and content creation \cite{Maharana_Hannan_Bansal_2022,Nazarieh_Feng_Awais_Wang_Kittler_2024}.
Typical paradigms include image captioning \cite{Daneshfar_Bartani_Lotfi_2024}, story generation \cite{Park_Yang_Jung_2023}, VQA 
 \cite{Zakari_Owusu_Qin_Wang_Lawal_He_2025}, and MMCW \cite{Li_Tang_Zhao_Nie_Wen_2022}. 
 % -- 简要介绍一下Visual storytelling任务
% Notably, MMCW exemplify the forefront of this domain.  
% MMCW generate artistically coherent textual output from a given user input.
% This field introduced VIST, which has inspired significant advancements.
% Various frameworks have emerged\cite{Kim_2016,Yu_Chung_Yun_Kim_Kim_2021} utilizing models based on convolutional neural networks (CNNs) combined with recurrent neural networks (RNNs), long shortterm memory networks (LSTMs), and Transformerbased architectures.
% Notably, MMCW exemplify the forefront of this domain.
% Recent studies utilize LLMs to integrate multi-modal information.
Generally, these tasks are confronted with multi-modal semantic inconsistency \cite{Yu_Chung_Yun_Kim_Kim_2021}.
Among these, story generation and MMCW will additionally encounter creative challenges.

% Specifically, several studies have validated in MMG.
% -- 写一些前人的工作
StoryGPT-V \cite{Shen_Elhoseiny_2023} leverages the latent diffusion and LLM to produce images grounded on given story descriptions.
SEED-Story \cite{Yang_Ge_Li_Chen_Ge_Shan_Chen_2024} has succeeded in generating up to 25 consecutive scene images and creating relatively long content via MLLMs.
%  -- 大家都用LLM，LLMs的成本高，让任务具有不小的门槛
% Many studies are based on LLMs. Nevertheless, it is crucial to acknowledge that deployment costs as well as API call expenses are frequently disregarded during the adoption of these models.  Such oversight can lead to underestimating the overall algorithmic overhead.
% \subsection{Multi-modal Creative Writing}
% 现有的方法主要在意生成文本的连贯性，而忽略了语义一致性和创造性
% Besides, some existing approaches have emphasized narrative coherence at the expense of semantic consistency \cite{Fan_Lewis_Dauphin_2018} and creativity \cite{Rashkin_Celikyilmaz_Choi_Gao_2020,Tang_Lin_Huang_Guerin_Zhang_2022}.
% 有些人提出方法尝试保证语义一致性，但对数据和模型参数的规模敏感。
To alleviate semantic inconsistency in MMG, \cite{Yang_Xiao_Huang_Zhong_2025,Yang_Ge_Li_Chen_Ge_Shan_Chen_2024} apply MLLMs, which are sensitive to the quantity of data and model parameters, in their method.
% 有人通过图文对齐方式解决这个问题，但模态不灵活
For the same purpose, \cite{Shen_Elhoseiny_2023,Wang_Zhao_Chen_Chen_Zheng_Shen_2024} proceed image-text semantic alignment with rigid interaction. 
% 为了提高创造性，但生成内容受限于训练数据分布
% \cite{Park_Yang_Jung_2023} tries to promote creativity by simply supplementing the training data, which is lacking of performance guarantee and poor in generalization.
% However, generating content beyond the boundaries of the training distribution continues to challenges.
% 又有人用llm干，通过改temperture来做，但是随机性不等于创造性
The DOC \cite{Yang_Klein_Peng_Tian_2023} also leverage LLMs to create story, which contains a detailed outliner and a controller.
It leverages the randomness in LM and set elevated temperature parameters for more creative result.
% 添加的baseline内容
% Additionally, methods excelling in caption and VQA tasks, such as mm-cot \cite{zhang2023multicot}, LaDic \cite{wang2024ladic}, and mPLUG-Owl \cite{ye2023mplugowl}, have been attempted for MMCW. 
LaDic \cite{wang2024ladic} propose a diffusion-based method for image caption.
The mm-cot \cite{zhang2023multicot} constructs a two-stage pipeline for MMG, enhancing multi-modal integration and logical reasoning.
mPLUG-Owl \cite{ye2023mplugowl} propose a training paradigm that equips LLMs with multi-modal abilities through modularized learning of foundation LLM, a visual knowledge module, and a visual abstractor module.

Despite the widespread of LLMs and MLLMs in MMG, the substantial computation and data demand, and rigid interactive patterns become non-negligible challenges in practice.
% Besides, the cost of API calls is often disregarded in the adoption of those methods. 
% Such an oversight can lead to an underestimation of the overall algorithmic overhead.
Moreover, although some aforementioned approaches demonstrate the capability to multi-modal generation, it is still lacking of exploration about alleviating multi-moda semantic inconsistency and creativity deficit.

\section{Method}
\begin{figure*}[htbp]
    \centering
    \includegraphics[width=1\textwidth]{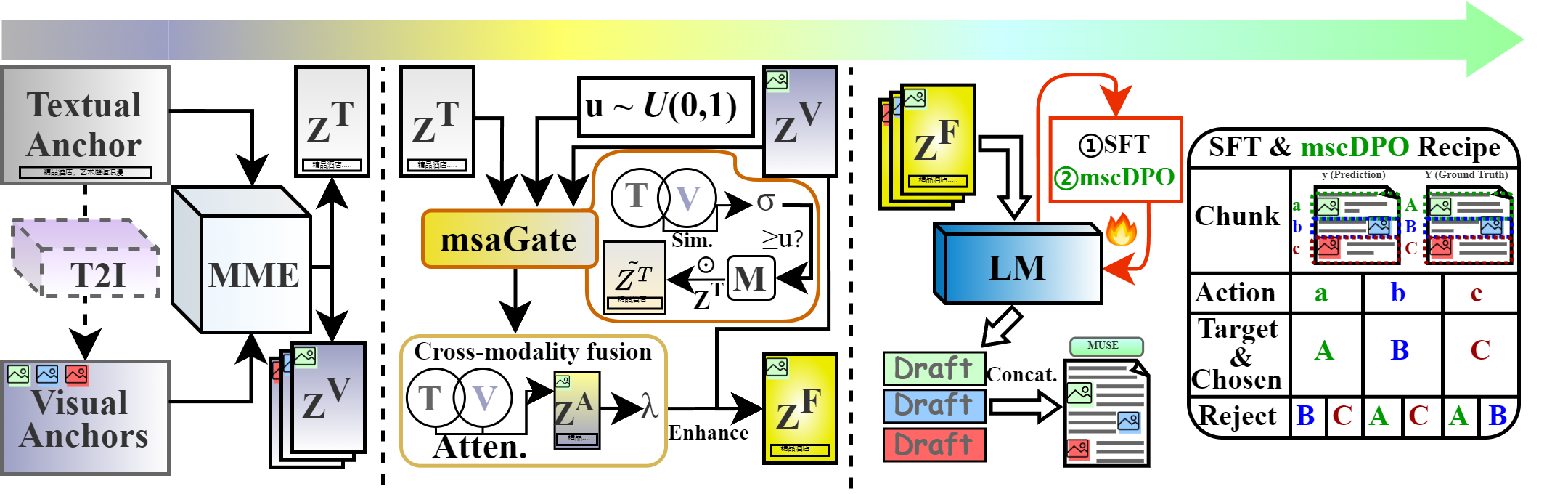}
    \caption{
    An overview of FlexMUSE, a flexible and efficient MMCW framework equipped with msaGate, cross-modality fusion module, and mscDPO design.
    % An overview of FlexMUSE.
    % The textual and visual input are extracted by multi-modal encoder.
    % Then msaGate and cross-modality fusion module constrain and enhance semantic knowledge.
    % Finally, LM trained with SFT and mscDPO is equipped to generate vivid MUSE.
    % The msaGate probabilistically masks the information from textual input to constrain textual cues that are similar to the visual semantics, focusing the LMs refer to the information from images.
    % The cross-modality fusion module is implemented by a multi-modal attention mechanism to seek common ground semantic knowledge of heterogeneous modalities, which is then augmented with feature augmentation to enhance semantics.
    % The mscDPO is designed to mitigate the quality degradation caused by SFT with respect to semantic consistency.
    } 
    \label{fig:FlexMUSE}
\end{figure*}
Equipped with optional T2I module (e.g., Stable Diffusion \cite{SD}), FlexMUSE, as shown in Fig.\ref{fig:FlexMUSE}, is able to enter text and images for MMG, or use only text (e.g., title or topic) to first create illustrations and then MUSE for flexible interaction.
Textual and visual inputs are treated as anchors and first extracted by multi-modal encoder (MME).
Then, the msaGate and cross-modality fusion module constrain and enhance semantic knowledge.
The processed information is then fed to the  downstream LMs.
Finally, LM trained with SFT and mscDPO produces the vivid MUSE.
The details are expanded below.

\subsection{Definition and initialization}
For the case that only the textual anchor $T$ is given, FlexMUSE generates $n$ semantic related but diversified images as the visual anchor $\mathbf{V}=\{V_1, \cdots, V_n \}$ by T2I module.
When both text and images are given, it can be considered as a special case that skipping image generation.
Supposed that the FlexMUSE once generate/achieve $n$ images, the textual input will also be repeat $n$ times and grouped with each image $V_i$ to construct an input set $I=\{(T,V_1), \cdots, (T,V_n)\}$.
Concluded the all parameter of any MMCW method $F$ as $\theta$, then the output $O= F(I,\theta)$ can be formulated as the combination of textual tokens and images, or the combination of text and corresponding visual placeholders.
% Since the training process of the visual generator is not our main concern, the MMCW can be regarded as a special language generation task.
For the sake of clarity, we omit the subscript of $V_i$ and use $V$ to represent one figure in all following statements.
% 符号定义
Initially, two modalities input will be first extracted by a MME $\phi_{*}^{*}$ (such as CLIP), $Z^T = \phi_{MME}^T(T)$, $Z^V = \phi_{MME}^T(V)$, where $ Z^* \in \mathbb{R}^{1 \times D}$.
Then, the $Z^T$ and $Z^V$ are injected into the subsequent generation.

\subsection{Modalities semantic alignment gate}
% 关于msaGate
% 然而，在实践中直接向LMs注入这两类信息容易使生成的内容陷入图文语义不一致或创意不足的情况。
With multi-modal injection, downstream modules are prone to suffer from inter-modal semantic inconsistencies in practice.
% 受信息论的启发，我们提出了msaGate，一种简单且高效的门控机制，用于约束文本信息的输入。
Inspired by information theory, we propose a simple but effective way to constrain the knowledge from textual anchor, named modalities semantic alignment gating mechanism (msaGate) as a countermeasure.
% 介绍msaGate思想：尝试阻止冗余的信息流入LM，以增加系统的稳定性,提高生成质量。
It blocks the redundant information to downstream modules by masking textual inputs based on the semantic similarity between modalities, for encouraging downstream module to refer more about visual anchors.

% 详细介绍msaGate
Specifically, for each input $(T,V)$, a masking threshold $u \sim U(0,1)$ is firstly sampled from a standard uniform distribution.
Then $\sigma$ records the semantic relevance within modalities by any semantic similarity function.
We apply the cosine function of the latent representations, as Eq \ref{eq:sigma}, to measure this similarity in our practices for the sake of generality and practicability:
\begin{align}
    \label{eq:sigma}
    \begin{aligned}
    \sigma &:= SemanticSim(T,V) = \frac{|Z^T \cdot Z^V|}{||Z^T||\cdot||Z^V||}.
    \end{aligned}
\end{align}
The msaGate filters out the feature $\tilde{Z^T}$ in $\mathbb{R}^{1 \times D}$ as $\tilde{Z^T} = M \odot Z^T$, where the $M\in$ $\mathbb{R}^{1 \times D}$ is a null vector when $\sigma \geq u$, or all-ones vector otherwise.

As the semantic gap narrows, $\sigma$ increases, reflecting more redundant information might exist within input.
Then, the msaGate becomes more inclined to mask such textual feature.
% 值得注意的是，因为msaGate是基于概率进行设计的，因此我们可以简单的通过拓展一个超参数或可训练参数，使该设计退化为恒等设计。
It is worth noting that, since the masking process is probabilistic, msaGate not only allows flexible tuning by adjusting the sampling distributions according to the data characteristics on inference, but also introduces controlled randomness that helps prevent overfitting to a fixed threshold on training, improves robustness to distribution shifts \cite{Ma_Li_Shi_Chen_2025}, and encourages better generalization across diverse scenarios.

\subsection{Cross-modality fusion}
% 介绍关于模态融合与语义增强
In MMG, to mine common knowledge across modalities while preserving modality-specific diversity, we design a cross-modality fusion module (Eq.\ref{eq:ZA}–\ref{eq:ZF}).
It first fuses heterogeneous input anchors via an attention-based fusion (Eq.\ref{eq:ZA}), then computes adaptive fusion weights (Eq.\ref{eq:lambda}), and finally applies feature augmentation to amplify modality-specific semantics while retaining shared information (Eq.\ref{eq:ZF}). 
This design enables the model to capture shared semantics while preventing modality-specific signals from being diluted during fusion.

% 关于Cross-modality fusion
% 先说融合
Specifically, to fuse the knowledge from multi-modal input, we use the idea in \cite{vaswani2023attentionneed} for mapping the visual query $Q=W^Q\odot{Z^V}$ and textual key-value pairs $K=W^K\odot\tilde{Z^T}$, $V=W^V\odot\tilde{Z^T}$ to an aggregated feature $Z^A \in \mathbb{R}^{1 \times D}$, where all the $W^* \in \mathbb{R} ^ {1 \times D}$ are trainable parameters:
% 要补充说，我们这里是在用图片来查文本，得出来的Za是关于文本的信息
\begin{align}
    \label{eq:ZA}
    Z^A =Attention(Q,K,V)=softmax(\frac{QK^T}{\sqrt{d_K}})V.
\end{align}
$Z^A$ tends to emphasize the visual information, which are close to the textual input in semantic, but mights overlook some textual knowledge.
For mining these neglected semantic knowledge, a trainable correlation $\lambda$ is proposed (Eq.\ref{eq:lambda}), where $\psi$ is the element-wise sigmoid function in our practice.
Then a semantic enhancement via feature augmentation is processing as Eq.\ref{eq:ZF} for amplify the specific while retaining the shared knowledge:
\begin{equation}
    \label{eq:lambda}
     \lambda = \psi(W^{\lambda} \cdot (Z^A)^{\mathsf{T}} \cdot Z^V) \in \mathbb{R}^{1 \times D}
\end{equation}
\begin{align}
    \label{eq:ZF}
    Z^F = (1-\lambda) \odot Z^A + \lambda \odot Z^V.
\end{align}

\subsection{Modalities semantic creative DPO}
% Unlike ordinary generative tracks, MMCW contains a variety of writing styles and very abstract semantics.
After training by SFT manner, the generated results of LM are prone to suffer from creative degradation.
To enhance creativity, carefully designed data or labels under the constraint of topic unification over paragraphs are necessary to be additional injected for SFT, which is low efficiency and difficult. 
Mimicking the human habit, RLHF, and DPO, the modality semantic creative direct preference optimization (mscDPO) is proposed to further tuning the LMs.
Equipped with mscDPO, LM can obtain two aspects of semantic supervision from positive (chosen) and negative (rejected) targets.
The details are described below.

% Specifically, to achieve outstanding writing quality, humans hold the writing preferences that the semantics of each illustration within a high-quality MUSE is strongly correlated with the adjacent text.
% Each visual anchor can be used as an illustration in the generated output and grouped with neighboring text generation to construct a chunk, considered as a semantic unit.
% We consider that, for a popular MUSE, the textual semantics from various chunks should be close to the textual anchor but not exactly the same.
% To achieve outstanding writing quality, humans hold the preferences that the semantics of each illustration is strongly correlated with the adjacent text while slightly different to.

For MMCW, it is reasonable to assume that each image $V$ in $I$ should close to the corresponding generated textual content in semantic. 
Thus, we can mark each image $V$ and the corresponding generated text as a semantic unit and split the entire MUSE to several chunk with this recipe which is applicable to both generating results and corresponding references (also called ground truth).

Consider that each $V$ is expanding from the same textual anchor $T$, thus the semantic gap between each chunk is controllable.
% Therefore, for each result $o_i = F((T,V_i), \theta)$, not only the corresponding ground truth chunk is naturally suitable for setting as chosen sample, the remaining chunks about other result are quite suitable for setting as rejected sample, which are diversify but related.
Based on this view, mscDPO adjusts the choosing strategy in conventional DPO (each input question is paired with one chosen answer and one rejected answer).
In mscDPO, for each generated result, the chosen answer is set as the corresponding chunk in corresponding references, while the remaining chunks in the same references are all set as rejected samples, which are naturally diversified but related in semantic.

% rejected answer are expanded, which  

% about chosen answer 

% The mscDPO adjusts the strategy in choosing the rejected samples which select any chunk for positive target and the other chunk for negative target within one generated illustrated article, simulating the one-to-many relationship.

% 这部分关于数据集数量的具体数字需要再做修改 
\section{ArtMUSE Datasets}
% %艺术性数据集少，所以我们提出ArtMUSE
% Despite the promising development in MMG, the manually calibrated corpuses for MMCW with text-image pairs are rare in the Chinese and even in the English community.
% To promote community development, we release a dataset for MMCW namely ArtMUSE, which contains with around 1k calibrated text-image pairs.  
% % 怎么来的
% The construction of ArtMUSE can be concluded within two steps, harvest and filtration.
% Firstly, we collected multifarious digital illustrated articles, especially those topics around architecture, design, and advertising.
% After collection, the automatic grammar checker and detection tool about resolution are leveraged to eliminate the low-quality content with obvious typo (less than three grammar errors within 20 tokens) and blurred images (over 128 $\times$ 128).
% Then the commons and reviews from 100 experts which are major in Arts, Architecture, Landscape Architecture, and Interior Design. 

% 为什么要公开ArtMuse
Despite advancements in MMG, there is still scarcity in manually curated corpora for MMCW with text-image pairs in both Chinese and English contexts. 
To support community development, we introduce a dataset for MMCW named ArtMUSE \footnote{The ArtMUSE will be declared in public after peer review for data privacy.}, which contains approximately 3,000 curated text-image pairs. 
% 怎么构建的，分两步，收集和过滤
% ArtMUSE 的创建包含两个主要阶段：数据收集和筛选校准。在数据收集阶段，我们针对艺术行业的微信官方账号进行收集，积累了大量的数字文本和图像内容，特别是在建筑、设计和广告领域。筛选和校准阶段使用自动广告过滤工具和图像分辨率检测工具进行初步筛选，剔除明显的广告内容和分辨率低于 128×128 像素的图像。随后，由 30 名艺术、建筑、景观建筑和室内设计方面的专家共同审查和校准数据，从而形成高质量的数据集。
The construct of ArtMUSE dataset involves two main phases: (1) data collection and (2) screening calibration.  
For the collection, we filter and gather the text and images content in social media from the architecture, design, and advertising sectors within the art and architectural design industries. 
During the calibration, we utilizes automated  filtering and image resolution detection tools to remove obvious textual advertisement and low resolutions images ($\leq$ 128$\times$128 px).  
Subsequently, a team of 30 experts major in art, architecture, and landscape architecture work together to calibrate and form the final result.

% 数据集的特点是什么，图文一致，包含创意文本
% ArtMUSE 数据集具有若干显著特点。它包含 263 篇文章、4832 张图片以及 15235 个与中国艺术相关的字符。该数据集确保了文本描述与图片的语义内容之间有紧密的对应关系。此外，它还包含了大量富有创意的文字，为多媒体内容创作研究提供了宝贵的素材。值得注意的是，ArtMUSE 提供的高分辨率图片尺寸超过 1024×1024 像素，而其他多模态生成（MMG）数据集通常提供的图片分辨率仅为 128×128 像素。
The ArtMUSE dataset ensures a semantic alignment between text and images in each MUSE. 
Additionally, it includes a substantial amount of creative text, offering valuable test-bed for MMCW. 
Notably, contrasts with other MMG datasets that typically offer images at a resolution of 128$\times$128 pixels, the resolution of images in ArtMUSE are obviously higher, with 1024$\times$1024 pixels.

% Compared with existing datasets that utilize simple and descriptive language, ArtMUSE is more artistic and creative, which is available at \url{https://anonymous_website_with_FlexMUSE_and_ArtMUSE.com}

\section{Experiments}

\subsection{Experimental settings}
    \paragraph{Baselines.}
    % We compare our method against the following baselines, selected for two primary reasons: (1) they represent the state of the art (SOTA) in their respective domains, enabling fair and meaningful comparisons with top-performing approaches; and (2) they are all open-source, ensuring reproducibility and transparent evaluation.

    %     The MMCW task requires generating creative text of considerable length, so we choose the story generation method DOC \cite{Yang_Tian_Peng_Klein_2022} and the online LLMs with visual input support GPT-4o \cite{openai2024gpt4ocard}, GLM-4V \cite{glm2024chatglmfamilylargelanguage}, Qwen2.5-vl \cite{bai2025qwen25vltechnicalreport} as our baselines.

    In this work, we selected baselines for two primary conditions: (1) these methods represent the state-of-the-art (SOTA) in respective domains, enabling fair and meaningful comparisons; (2) they are all open-sourced, ensuring reproducibility and transparent evaluation.
    We set the aftermentioned method as baselines: (1) DOC \cite{Yang_Tian_Peng_Klein_2022},  the story generation method; (2) mm-cot, the multi-modal reasoning method \cite{zhang2023multicot}; (3) LaDic \cite{wang2024ladic}, the vision-language model; (4) mPLUG-Owl \cite{ye2023mplugowl}, the visual question answer framework; (5) GPT-4o \cite{openai2024gpt4ocard}, GLM-4V \cite{GLM_2024}, and Qwen2.5-vl \cite{bai2025qwen25vltechnicalreport}, the online LLMs enable visual input.
    \paragraph{Measurements.}
    In order to measure the quality of the generated results as comprehensively as possible, we set two types of multidimensional criteria in our setting, namely automatic evaluation and LLMs judgment.
    
\begin{figure}[htbp]
    \centering
    \includegraphics[width=0.95\linewidth]{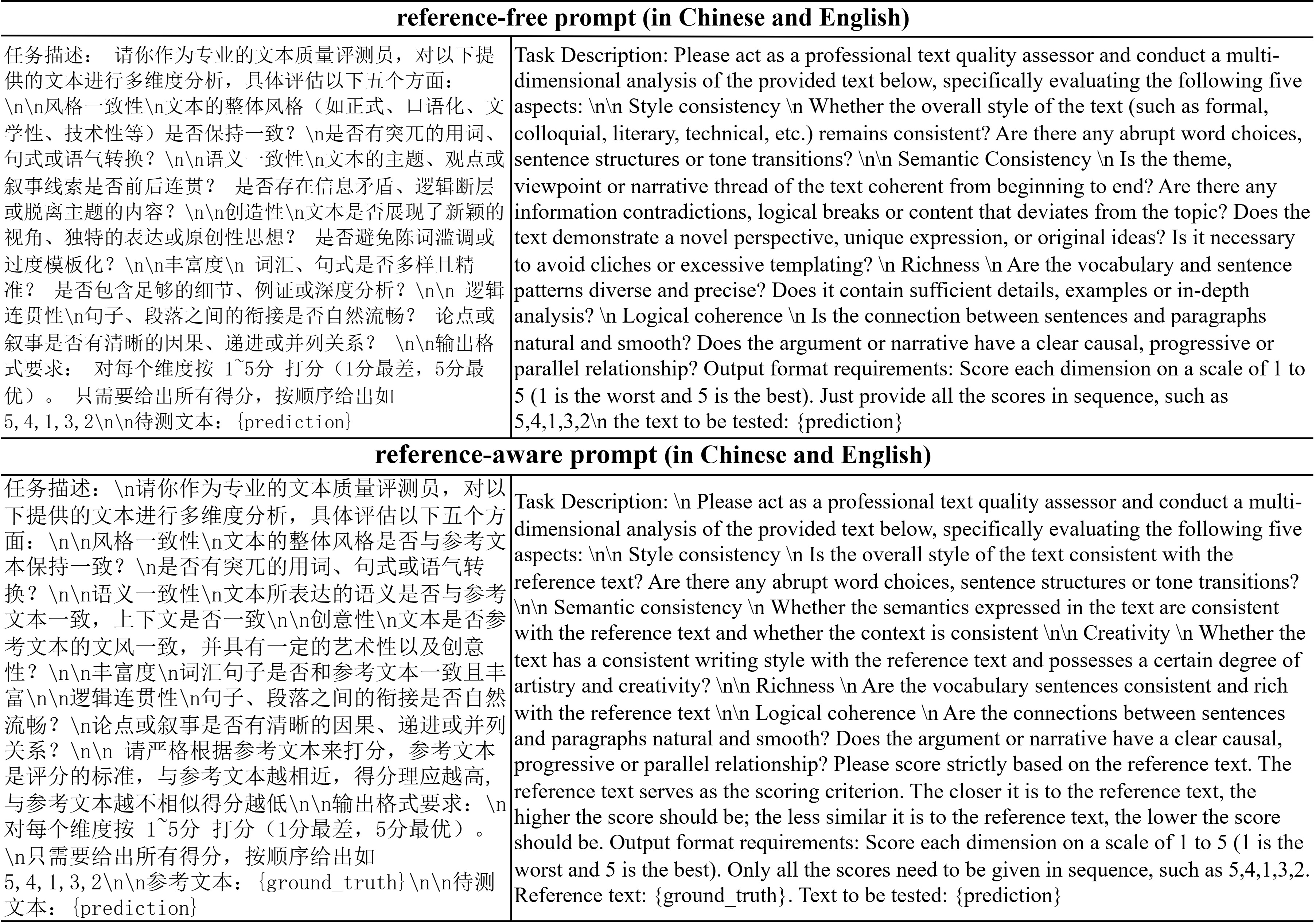}
    \caption{Textual prompt leveraged in LLM evaluation.}
    \label{fig:prompt}
    \end{figure}

    For the automatic evaluation, we use ROUGE-1, ROUGE-2, ROUGE-L and BertScore following \cite{Xue_Qian_Fang_Xu_2022,Yang_Xiao_Huang_Zhong_2025} to measure the similarity between the generated textual results and the ground-truth. 
    Meanwhile, since the automatic metrics may not comprehensive enough to measure the quality, we leverage LLMs to judge the result from five perspectives and propose five groups of original metric.
    For each group, we propose two prompts, as shown in Fig. \ref{fig:prompt}, for the reference-free evaluation and reference-aware evaluation, respectively.
    For reference-free evaluation, scored from 0-5, we append the superscript $f$ to the corresponding criterial abbreviation.
    Similarly, for the reference-aware evaluation, also scored from 0-5, the superscript $a$ are used.
    The details are as follows:
    \begin{itemize}
        \item Style Consistency ($SC^f, SC^a$): Measuring whether the tone (formal, informal, literary, technical) matches and identifies any language shifts that feel out of place.
        
        \item Context Consistency ($CC^f, CC^a$): Measuring the degree of theme stays consistent, whether the ideas remain are consistent, the degree of narrative flows smoothly, whether logical errors or off-topic parts existed. 
        
        \item Creativity ($CV^f, CV^a$): Evaluating how well the text presents new ideas, unique expressions, and original thoughts while avoiding clichés.
        
        \item Richness ($RN^f, RN^a$): This looks at word variety, sentence complexity, and detail depth.
        
        \item Coherence($CO^f, CO^a$): Examines how sentences and paragraphs connect and how strong their semantic relationships are.
    \end{itemize}

\paragraph{Details.}
    Denoted $\eta$ as the learning rate.
    For the multimodal entering, we utilized the pre-trained CLIP \cite{yang2023chineseclipcontrastivevisionlanguage} to extract features from both textual and visual inputs.     
    The T5 decoder is initialized from langboat/mengzi-t5-base for ArtMUSE, acting as LM in our experiment.
    During SFT, FlexMUSE trains 500K iterations with $\eta=8e-5$, verified from \{5e-5, 6e-5, 7e-5, 8e-5, 9e-5\}. 
    The mscDPO executes 200K iterations with $\eta=1e-5$, verified from \{5e-6, 6e-6, 7e-6, 8e-6, 9e-6, 1e-5, 2e-5\}.     
    All experiments are executed on NVIDIA RTX 4090 GPU.
\subsection{Experimental Results}
\begin{figure}[htbp]
            \centering
            \includegraphics[width=0.8\linewidth]{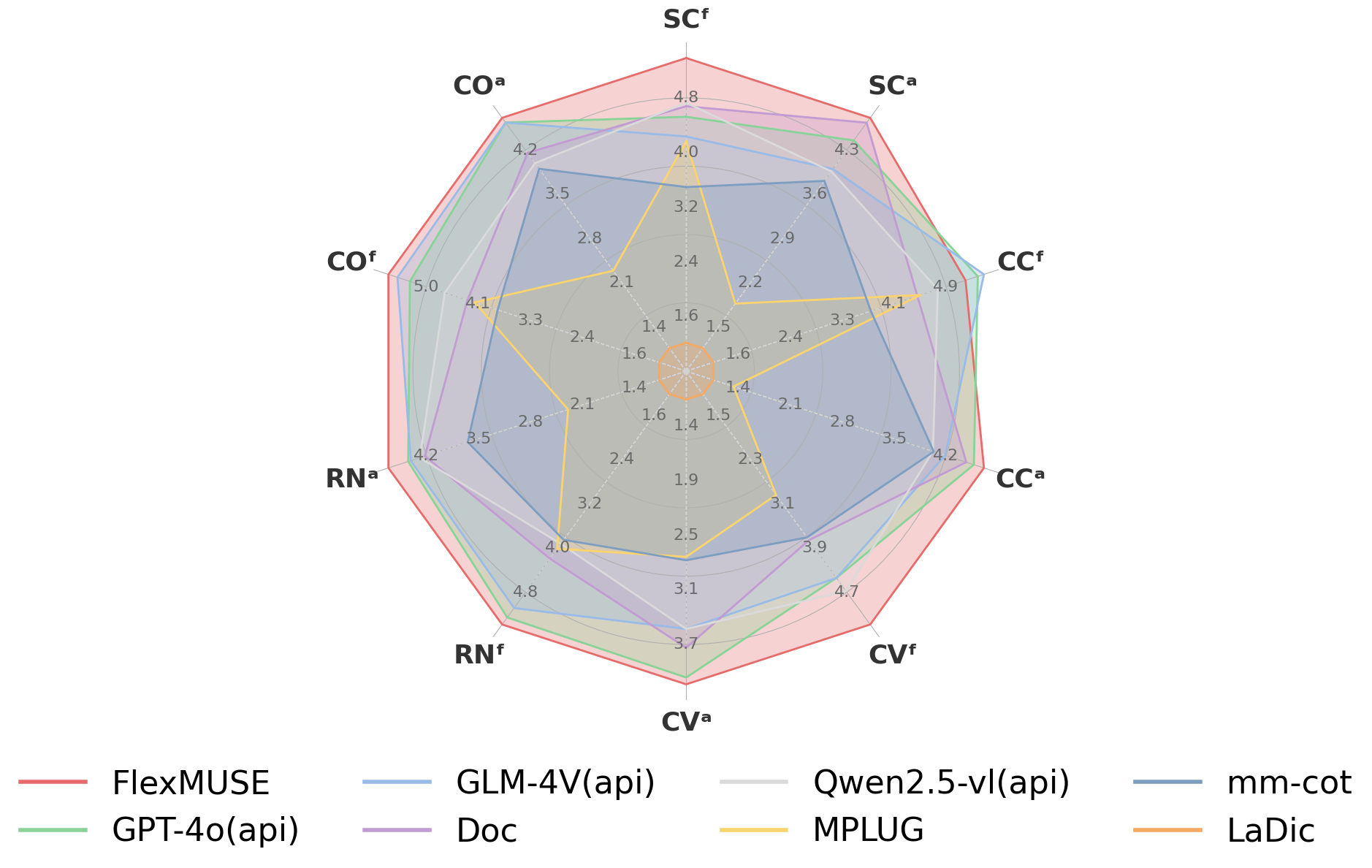}
            \caption{
            LLM evaluation of different methods on ArtMUSE. The FlexMUSE $SC^f$, $SC^a$, $CC^f$, $CC^a$, $CV^f$, $CV^a$, $RN^f$, $RN^a$, $CO^f$, $CO^a$ scores were $4.5$, $3.98$, $4.31$, $3.90$, $4.4$, $3.48$, $4.47$, $3.7$, $4.63$, $3.93$, respectively.
            }
            \label{fig:LLMscore1}
        \end{figure}

    \paragraph{Automatic evaluation.}
        \begin{table}[h]
              \centering
              \caption{
              The results of different methods with 3 automatic evaluation metrics.
              }
                \scalebox{0.9}{  
                    \begin{tabular}{lcccc}\toprule
                    
                    \text{ArtMUSE}& \text{ROUGE-1}& \text{ROUGE-2}& \text{ROUGE-L}& \text{BertScore}
\\\midrule
                    
                    mm-cot& 0.01 & 0.01 & 0.12 & 0.43 
\\
                    mPLUG-Owl& 0.13 & 0.02 & 0.14 & 0.63 
\\
                    LaDic& 0.12 & 0.00 & 0.12 & 0.55 
\\
                    DOC& 0.13 & 0.03 & 0.08 & 0.65 
\\
                    
                    GLM-4V& 0.21 & 0.06 & 0.09 & 0.66 
\\
                    
 Qwen2.5-vl& 0.17 & 0.04 & 0.09 &0.65 
\\
 GPT-4o& 0.19 & 0.05 & 0.09 &0.66 
\\\midrule

 \textbf{FlexMUSE}& \textbf{0.55} & \textbf{0.34} & \textbf{0.43} &\textbf{0.72} 
\\ \bottomrule
                    \end{tabular}%
                }
              \label{tab:automatic_score}%
            \end{table}%
        % As shown in Table \ref{tab:addlabel}, FlexMUSE shows promising performance within 4 evaluation metrics, where ROUGE-L and BertScore improve by 0.35 and 0.07 relative to DOC, the only open source method.
        % Meanwhile, compared with other LLM-based method,
        % FlexMUSE also have the upper hand.
        % % 除了CC都高，体现创造性和连贯性
        % Given that these four metrics are focused on character level, we use another set of metrics for further comparison.
        % \textcolor{blue}{As shown in Table \ref{tab:addlabel}, FlexMUSE achieves significant improvements across all four evaluation metrics. Specifically, our method reaches 0.55 in ROUGE-1, surpassing mm-cot by 0.54, 0.34 in ROUGE-2, exceeding LaDic by 0.34, 0.43 in ROUGE-L, outperforming DOC by 0.35, and 0.72 in BertScore, improving over mPLUG-Owl by 0.09. FlexMUSE consistently outperforms both open-source baselines and proprietary LLMs. Compared to the strongest open-source baseline (DOC), it achieves gains of 0.42 in ROUGE-1, 0.31 in ROUGE-2, 0.35 in ROUGE-L, and 0.07 in BertScore. Against the best-performing LLM (GLM-4V), FlexMUSE improves ROUGE-1 by 0.34, ROUGE-2 by 0.28, and BertScore by 0.06. LLM-based method, FlexMUSE also have the upper hand.Given that these four metrics are focused on character level, we use another set of metrics for further comparison.}
        Table \ref{tab:automatic_score} shows the results of the automatic evaluation, demonstrating that FlexMUSE surpasses all baseline models. It achieves scores of 0.55 in ROUGE-1, 0.34 in ROUGE-2, 0.43 in ROUGE-L, and 0.72 in BertScore metrics that significantly exceed those of both open-source methods, such as mm-cot (0.01 in ROUGE-1) and DOC (0.13 in ROUGE-1), as well as proprietary large language models (LLMs) like GLM-4V (0.21 in ROUGE-1) and GPT-4o (0.19 in ROUGE-1). 
        \paragraph{LLM-based evaluation.} As shown in Fig. \ref{fig:LLMscore1}, FlexMUSE leads in creativity ($CV^f, CV^a$) and coherence ($CO^f, CO^a$), producing text that is more novel and logically consistent than that generated by other methods. Although it slightly trails GLM-4V in no-reference context consistency ($CC^f$), it excels in reference-aware context consistency ($CC^a$), demonstrating stronger alignment with the ground-truth and outperforming most baselines in this critical metric.

\paragraph{Human evaluation.}
    
\begin{figure}[htbp]
        \centering
        \includegraphics[width=0.55\linewidth]{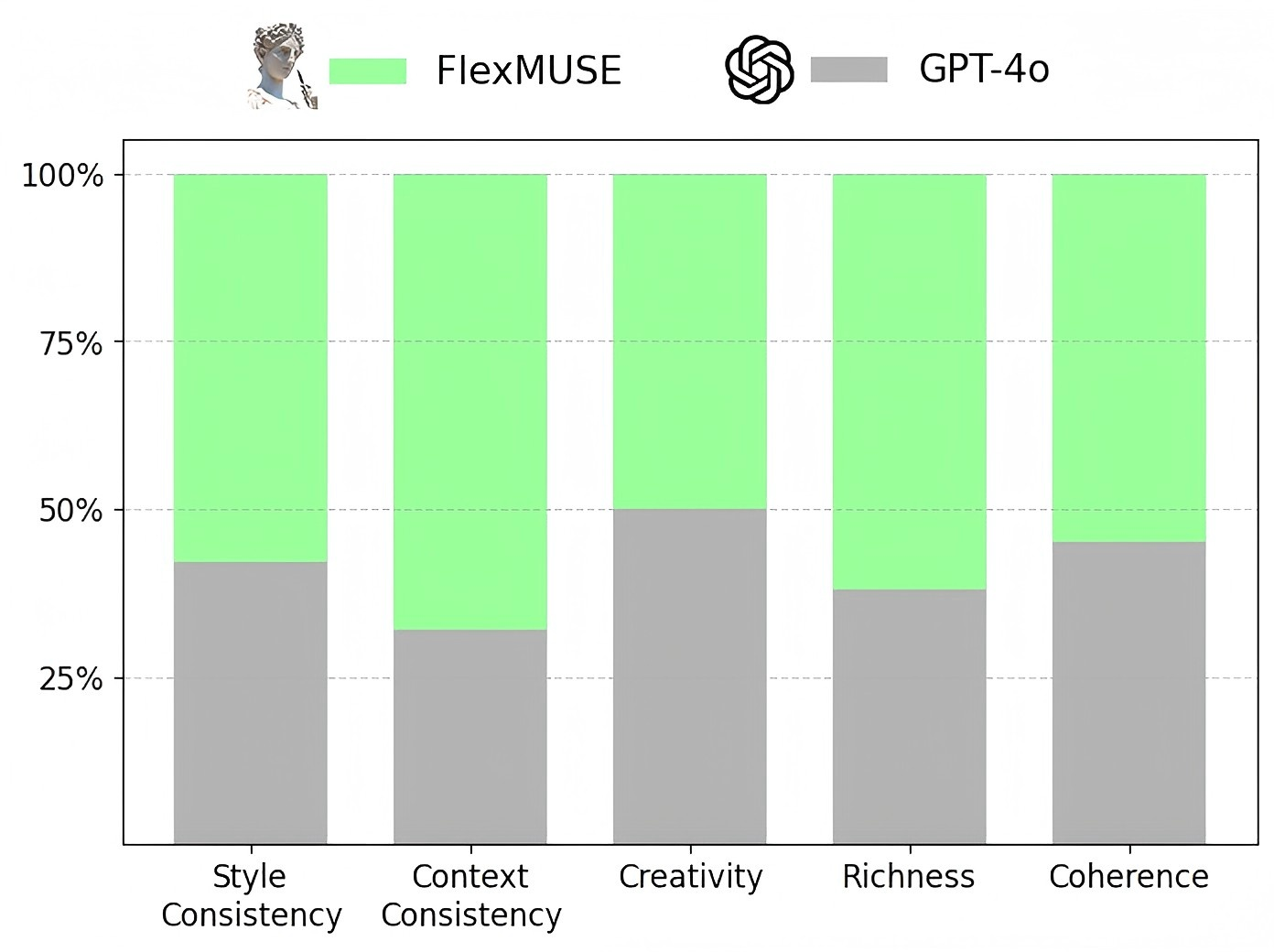}
        \caption{Human evaluation results on ArtMUSE include style consistency, context consistency, creativity, richness and coherence.}
        \label{fig:human}
    \end{figure}

    For more comprehensive comparison, we invite 50 experts which are major in Arts, Architecture, Landscape Architecture, and Interior Design to assess the quality of 100 articles generated by FlexMUSE or GPT-4o on ArtMUSE. 
    Specially, given a pair of articles generated by FlexMUSE and GPT-4o, experts are asked to vote which generated article is better include style consistency, context consistency, creativity, richness and coherence.
    Each pair is evaluated by 5 unique experts and we summarized the result in Fig. \ref{fig:human}.

\paragraph{Effectiveness.}

\begin{table}[htbp]
    \centering
    \caption{Effectiveness of individual modules or combinations in FlexMUSE evaluated by both reference-free and reference-aware metrics. FlexMUSE achieves the highest scores across all metrics, with significant improvements over SFT.}
    \scalebox{0.9}{ % 适当缩小以适应页面宽度
        \begin{tabular}{lp{0.82cm}p{0.82cm}p{0.82cm}p{0.82cm}p{0.82cm}
            p{0.82cm}p{0.82cm}p{0.82cm}p{0.82cm}p{0.82cm}}
        \toprule
        \textbf{Method} & \multicolumn{5}{c}{\textbf{Reference-free Metrics}} & \multicolumn{5}{c}{\textbf{Reference-aware Metrics}} \\
        \cmidrule(lr){2-6} \cmidrule(lr){7-11}
        & \textbf{$SC^f$} & \textbf{$CC^f$} & \textbf{$CV^f$} & \textbf{$RN^f$} & \textbf{$CO^f$} & \textbf{$SC^a$} & \textbf{$CC^a$} & \textbf{$CV^a$} & \textbf{$RN^a$} & \textbf{$CO^a$} \\
        \midrule
        SFT   & 3.87  & 3.95  & 3.34  & 3.50  & 3.55 & 3.50  & 3.67  & 2.96  & 3.50  & 3.51 \\
        +DPO  & 3.98  & 3.95  & 3.18  & 3.57  & 3.55 & 3.85  & 3.85  & 3.22  & 3.63  & 3.72 \\
        +mscDPO & 4.18  & 4.34  & 3.81  & 3.93  & 4.02 & 3.88  & 3.85  & 3.42  & 3.66  & 3.82 \\
        +msaGate & 4.12  & 3.84  & 3.75  & 3.72  & 3.65 & 3.92  & 3.78  & 3.45  & 3.64  & 3.87 \\
        +msaGate+DPO & 4.22  & 3.96  & 2.89  & 3.33  & 3.71 & 3.71  & 3.71  & 3.38  & 3.61  & 3.80 \\
        \midrule
        \textbf{FlexMUSE} & \textbf{4.5}  & \textbf{4.31} & \textbf{4.4} & \textbf{4.47} & \textbf{4.63} & \textbf{3.98} & \textbf{3.9} & \textbf{3.48} & \textbf{3.9} & \textbf{3.93} \\
        \bottomrule
        \end{tabular}%
    }
    \label{tab:Combined Ablation Studies}%
\end{table}

As shown in Table \ref{tab:Combined Ablation Studies}, standard DPO combined with SFT left most performance metrics largely unchanged, indicating its lack of significant effectiveness in the evaluated metrics. In contrast, the proposed mscDPO yielded improvements across all metrics, suggesting its superiority due to enhanced cross-modal semantic consistency, stimulation of creativity via vivid visual cues, and improved topic adherence. Integrating the msaGate mechanism caused minor drops in some metrics, yet overall performance remained robust, with$ (SC^f = 4.12), (SC^a = 3.98), (CO^f = 3.65), and (CO^a = 3.93)$. Finally, combining msaGate and mscDPO with SFT (FlexMUSE) significantly improved both reference-free and reference-aware metrics, demonstrating the effectiveness of mscDPO and msaGate in enhancing performance for automated MMCW.

    % % 讲纯DPO一般
    % As shown in Tables \ref{tab:Combined Ablation Studies}, when we applied the standard DPO method in conjunction with SFT, most performance scores remained nearly unchanged, indicating that DPO might not be significantly effective in the concerned measurement.
    % % 讲mscDPO好
    % In contrast, equipped with the proposed mscDPO, all result increase indicating that mscDPO might be more effective than the standard DPO because it ensures semantic consistency between modalities, ability to stimulate imagination from vivid visual knowledge for more creative writing and reasonably keep text content on topic.
    % % 讲msaGate好
    % With msaGate mechanism resulted in slight declines in some scores, yet the overall performance remained robust, with $SC^f, SC^a$ reaching $4.12$, $3.98$ and $CO^f, CO^a$ achieving $3.65$, $3.93$. 
    % % Since the msaGate is proposed to prevent redundant textual information from flowing into LM, it could focuse the downstream modules pay more attention to more creative visual features.

    % % to encourage downstream modules to focus on  visual features for creative.
    % % 加起来也好
    % Finally, when combining msaGate and mscDPO with SFT (FlexMUSE), both reference-free metrics and reference-aware metrics are significantly improved. 
    % The results demonstrate the effectiveness of the mscDPO and msaGate in enhance the performance for automatic MMCW. 
    
% Table generated by Excel2LaTeX from sheet '总表'

\paragraph{Memory usage comparison.}
\begin{figure}[htbp]
    \centering
    \begin{subfigure}[b]{0.45\textwidth}
        \centering
        \includegraphics[width=\linewidth]{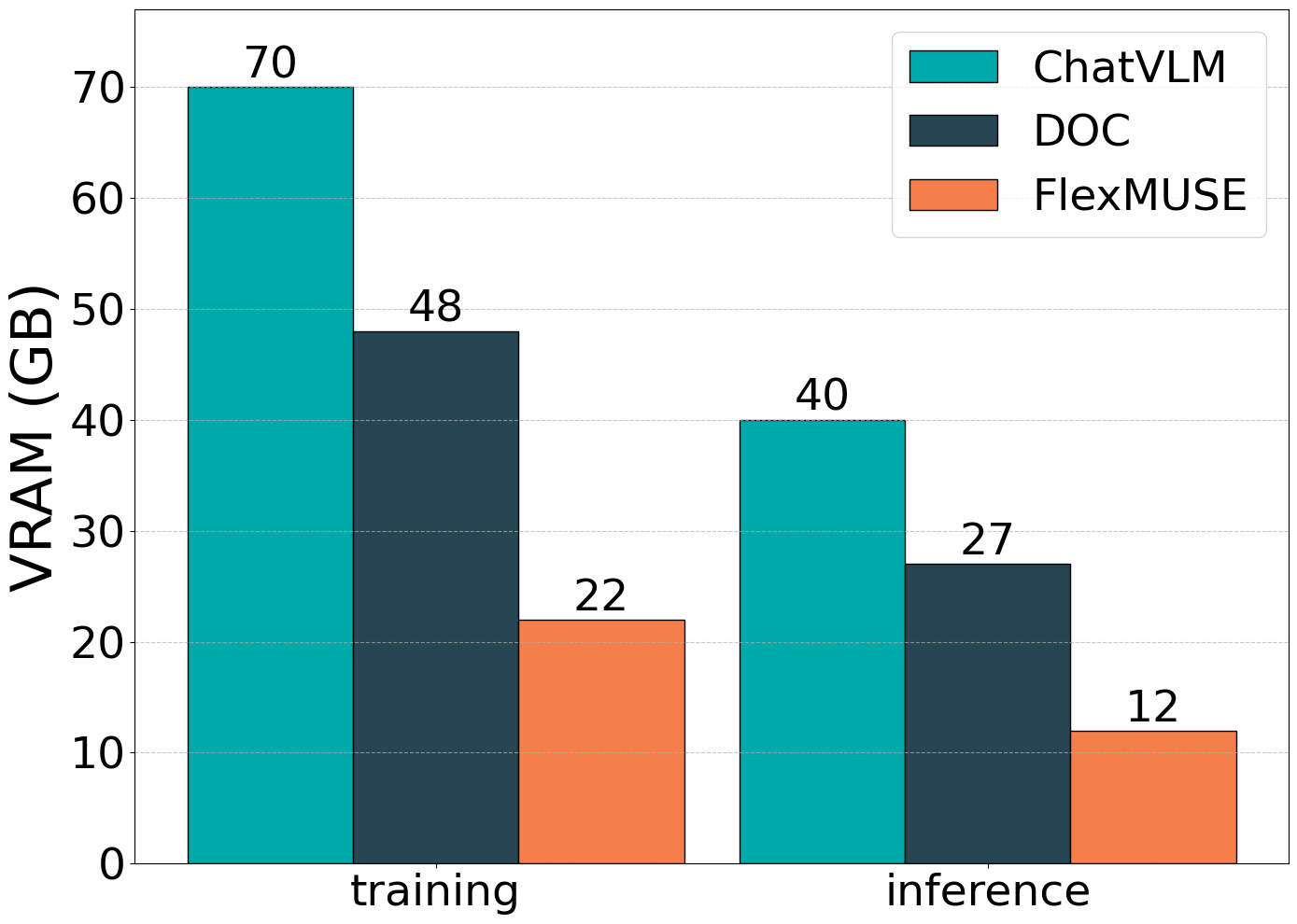}
        \caption{}
        \label{fig:VRAM}
    \end{subfigure}
    \hspace{0.01\textwidth}  % 减小子图间距
    \begin{subfigure}[b]{0.45\textwidth}
        \centering
        \includegraphics[width=\linewidth]{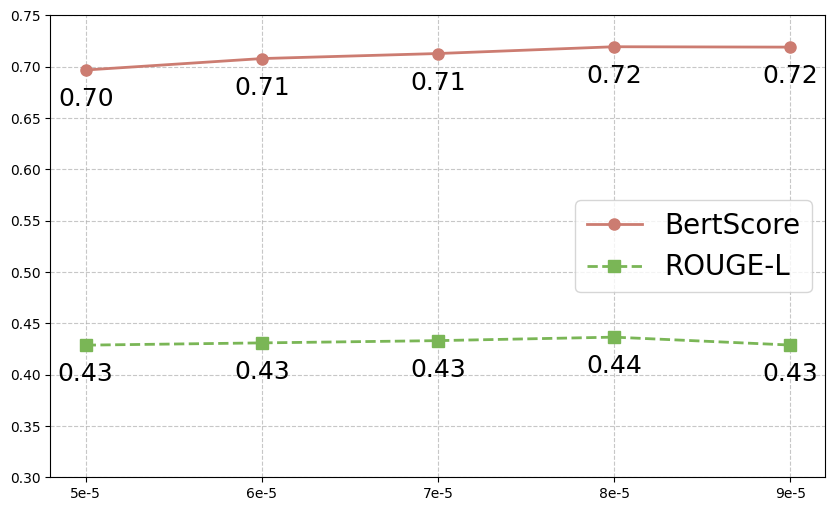}
        \caption{}
        \label{fig:Robusting}
    \end{subfigure}
    
    \caption{(a) Comparison about the peak VRAM usage of different methods in training/inference phase. (b) The performance of FlexMUSE with different learning rates (from 5e-5 to 9e-5) on ArtMUSE.}
    \label{fig:combined}
\end{figure}

    Beyond performance, computational efficiency is critical in practical applications. As Figure \ref{fig:VRAM} illustrates, FlexMUSE surpasses ChatVLM \cite{GLM_2024} and DOC in VRAM efficiency during both training and inference. Its streamlined architecture and targeted optimizations reduce overhead while maintaining or enhancing performance, facilitating access for users with limited resources. Unlike ChatVLM and DOC, which require powerful hardware or distributed setups due to high VRAM demands, FlexMUSE can be deployed on memory-constrained devices with minimal performance loss.

\paragraph{Robustness to hyperparameters.}
    We construct relevant experiments for the learning rate $\eta$ as shown in Fig. \ref{fig:Robusting}.
    For clarity, we only exhibit the evaluated result after SFT with different learning rates (from 5e-5 to 9e-5).
    % \begin{figure}[htbp]
    %     \centering
    %     \includegraphics[width=0.6\linewidth]{img/robust.png}
    %     \caption{The performance of FlexMUSE with different learning rates on ArtMUSE.}
    %     \label{fig:Robusting}
    % \end{figure}
    The experimental result demonstrates that the BertScore (red line) and ROUGE-L (green-line) metrics do not have obvious fluctuation with different setting, showing the insensitive of FlexMUSE about the training rate to some extent.

\section{Conclusion and limitation}
% In this work, we propose a flexible framework for multi-modal creative writing, namely FlexMUSE.
% To address the challenge of the semantic inconsistency between modalities, FlexMUSE is equipped with the msaGate to constrain the entering textual clue, facilitating the downstream module to refer to the visual knowledge.
% Besides, an attention-based cross-modality fusion module is involved in FlexMUSE to capture the correlation between input text and images for semantic enhancement.
% Leveraging ideas from the human writing habit and RLHF, mscDPO is proposed to address the quality degradation caused by SFT via extending the rejected samples in standard DPO for more diverse supervision from semantics.
% In addition, to advance community development, we propose a manually calibrated dataset for MMCW, namely ArtMUSE, which is collected from Chinese social media and has approximately 1k text-image pairs.

% With the help of several well-designed modules, FlexMUSE achieves promising performance in terms of consistency, creativity and coherence in illustrated articles.
% However, since open-source models for MMG are expensive to produce, comparisons between FlexMUSE and baselines with respect to different measurements are limited in this work and can be supplemented in the future to construct more comprehensive benchmarks, and the effect of hyperparameter in the msaGate module can be discussed further.

In this work, we propose a flexible framework for multimodal creative writing (MMCW), namely FlexMUSE. 
To address cross-modal semantic inconsistency, FlexMUSE integrates the msaGate to filter out the redundant knowledge from textual input. 
Additionally, an attention-based cross-modal fusion module is integrated into FlexMUSE to capture text-image input correlations for semantic enhancement. 
Drawing on human writing practices and RLHF, mscDPO is designed to mitigate SFT-induced quality degradation by expanding rejected samples in standard DPO, providing semantically diverse supervision. 
Furthermore, to support community progress, we introduce ArtMUSE, a manually calibrated MMCW dataset curated from Chinese social media comprising 3k text-image pairs.
Enabled by its well-designed modules, FlexMUSE achieves promising performance in consistency, creativity, and coherence for illustrated content. Due to the high cost of developing open-source MMG models, comparisons between FlexMUSE and baselines across diverse metrics are limited here; these will be expanded in future work to build more comprehensive benchmarks. 
% Additionally, the impact of hyperparameters in the msaGate module warrants further investigation.

\bibliographystyle{splncs04}
\bibliography{ref}

\end{document}